\RequirePackage{amsthm}
\documentclass[sn-mathphys-ay,iicol]{sn-jnl}% Math and Physical Sciences Numbered Reference Style 
\usepackage{graphicx}%
\usepackage{multirow}%
\usepackage{amsmath,amssymb,amsfonts}%
\usepackage{amsthm}%
\usepackage{mathrsfs}%
\usepackage[title]{appendix}%
\usepackage{xcolor}%
\usepackage{textcomp}%
\usepackage{manyfoot}%
\usepackage{booktabs}%
\usepackage{algorithm}%
\usepackage{algorithmicx}%
\usepackage{algpseudocode}%
\usepackage{listings}%
\usepackage{xcolor}
\usepackage{subfigure}
\usepackage{adjustbox}
\usepackage{lmodern}

\theoremstyle{thmstyleone}%
\theoremstyle{thmstyletwo}%

\theoremstyle{thmstylethree}%

\begin{document}

\title[Article Title]{
% LLMs are Expert in Classifying Images:
HPT++: Hierarchically Prompting Vision-Language Models 

with Multi-Granularity Knowledge Generation 

and Improved Structure Modeling
}


\author[1]{\fnm{Yubin} \sur{Wang}}\email{wangyubin2018@tongji.edu.cn}

\author[2]{\fnm{Xinyang} \sur{Jiang}}\email{xinyangjiang@microsoft.com}

\author[3]{\fnm{De} \sur{Cheng}}\email{dcheng@xidian.edu.cn}

\author[1]{\fnm{Wenli} \sur{Sun}}\email{2233055@tongji.edu.cn}

\author[2]{\fnm{Dongsheng} \sur{Li}}\email{dongsheng.li@microsoft.com}

\author*[1]{\fnm{Cairong} \sur{Zhao}}\email{zhaocairong@tongji.edu.cn}

\affil*[1]{\orgdiv{Department of Computer Science and Technology}, \orgname{Tongji University}, \orgaddress{\city{Shanghai}, \country{China}}}

\affil[2]{\orgname{Microsoft Research Asia}, \orgaddress{\city{Shanghai}, \country{China}}}

\affil[3]{\orgdiv{School of Telecommunications Engineering}, \orgname{Xidian University}, \orgaddress{\city{Xi’an}, \country{China}}}


\abstract{Prompt learning has become a prevalent strategy for adapting vision-language foundation models (VLMs) such as CLIP to downstream tasks. 
With the emergence of large language models (LLMs), recent studies have explored the potential of using category-related descriptions to enhance prompt effectiveness. 
However, conventional descriptions lack explicit structured information necessary to represent the interconnections among key elements like entities or attributes with relation to a particular category. 
Since existing prompt tuning methods give little consideration to managing structured knowledge, this paper advocates leveraging LLMs to construct a graph for each description to prioritize such structured knowledge. 
Consequently, we propose a novel approach called Hierarchical Prompt Tuning (HPT), enabling simultaneous modeling of both structured and conventional linguistic knowledge. 
Specifically, we introduce a relationship-guided attention module to capture pair-wise associations among entities and attributes for low-level prompt learning. 
In addition, by incorporating high-level and global-level prompts modeling overall semantics, the proposed hierarchical structure forges cross-level interlinks and empowers the model to handle more complex and long-term relationships. 
Finally, by enhancing multi-granularity knowledge generation, redesigning the relationship-driven attention re-weighting module, and incorporating consistent constraints on the hierarchical text encoder, we propose HPT++, which further improves the performance of HPT. Our experiments are conducted across a wide range of evaluation settings, including base-to-new generalization, cross-dataset evaluation, and domain generalization. Extensive results and ablation studies demonstrate the effectiveness of our methods, which consistently outperform existing SOTA methods.}




\maketitle

\section{Introduction}\label{sec1}
\begin{figure}[t]
    \centering
    \center{\includegraphics[width=7.6cm]{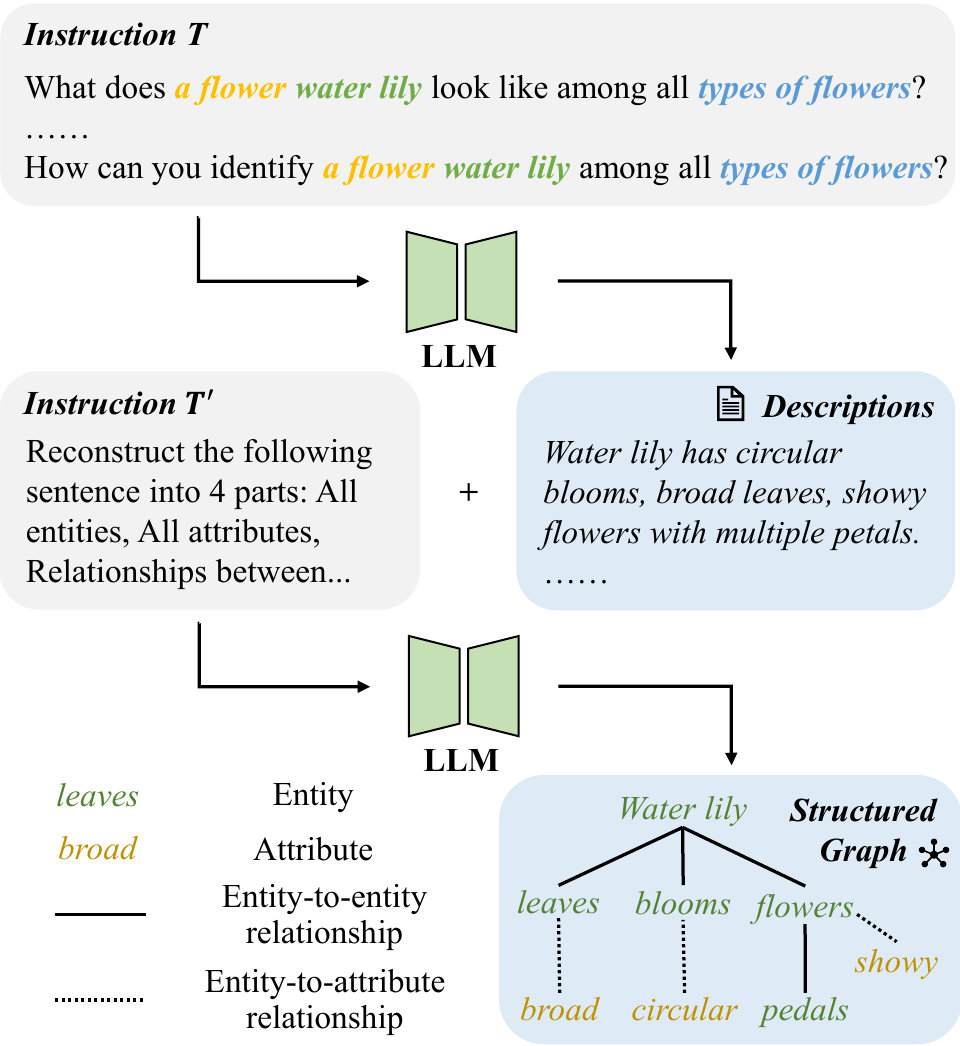}}
    \caption{We input a few hand-written instructions into LLMs to generate human-like category-related descriptions along with structured graphs based on each description. \protect\label{fig:intro}}
\end{figure} 
Vision-language foundation models (VLMs)~\citep{radford2021learning, jia2021scaling} have significantly advanced in learning transferable representations.
To effectively explore the potential of them, prompt tuning methods~\citep{zhou2022learning, zhou2022conditional, khattak2023maple} learn continuous vectors, known as prompt vectors, and incorporate them into the input space, thereby enhancing the representation capability of the pre-trained network. 
However, when faced with ambiguous category names, models often struggle to accurately interpret the corresponding visual concepts, resulting in sub-optimal performance. 
Thus, using category names as text input without the assistance of more linguistic knowledge is an unsatisfactory choice. 
Recent methods~\citep{zhang2023prompt, pratt2022does, menon2022visual} address this issue by employing large language models (LLMs)~\citep{brown2020language, llama3modelcard}. 
These methods use hand-written templates to generate human-like texts enriched with linguistic knowledge, thereby facilitating few-shot visual recognition.

In this paper, we propose a novel approach that enhances natural linguistic descriptions with structured knowledge representations. 
We assert that structured knowledge is essential for prompt tuning. 
Specifically, the descriptions of a category with unstructured knowledge consist of some key elements, such as entities and attributes, which define that category. 
For example, the category 'water lily' is defined by entities such as 'leaves', 'blooms', and 'flowers', each linked to specific attributes. 
Following related work on knowledge graphs~\citep{tay2017multi, zhang2021ma}, we represent these key elements, and their correlations as a graph for semantic understanding. 
This graph-based representation offers a more organized way to present information, enhancing data comprehension. 
It also facilitates the discovery of implicit connections that may not be evident in original descriptions. 
In this work, we leverage existing large language models to extract structured information from vanilla descriptions. 
Given a specific category, we feed handcrafted instructions into LLMs to generate human-like descriptions and structured relationships within them, including key elements and their interrelationships, as shown in Figure~\ref{fig:intro}.

However, existing prompt tuning methods are inadequate for explicitly modeling such structured knowledge represented as a graph. 
To address this issue, we propose \textbf{H}ierarchical \textbf{P}rompt \textbf{T}uning (HPT) to incorporate both structured and conventional linguistic knowledge from LLMs, enhancing prompt effectiveness hierarchically. 
To model complex structured information, HPT learns hierarchical prompts at different semantic levels. 
Specifically, HPT contains low-level prompts representing relationships among key elements of the category, high-level prompts with implicit category-related semantics derived from descriptions, and global-level prompts with task- or domain-specific knowledge shared across categories.

We introduce a relationship-guided attention module to leverage and model LLM-generated pair-wise correspondences among entities and attributes, where learnable attention-based matrices are integrated into the text encoder.
Furthermore, cross-level self-attention is adopted to model relationships between prompts from different levels to handle more complex and long-term relationships not fully exploited by LLMs. It effectively overcomes the limitations caused by relying solely on modeling low-level tokens and allowing for a more comprehensive understanding of the category.

Our prompts are trained under a dual-path asymmetric framework~\citep{zhao2024learning}, where prompted image encoder and text encoder are learned separately by aligning their output with frozen encoders from the other modality. By replacing the vanilla-prompted text encoder, which learns only category-agnostic prompts, with a novel hierarchical prompted text encoder, text representations align better with corresponding visual concepts, resulting in superior recognition performance. 

The contributions of HPT are summarized as follows. 
1) We raise the consideration that it is crucial to extract and leverage structured knowledge from descriptions to assist learning prompts. 
Thus, we firstly leverage large language models to generate category-related descriptions along with corresponding structured relationships. 
2) We propose hierarchical prompt tuning for simultaneously modeling both structured and conventional linguistic knowledge. 
By incorporating both forms of knowledge, we can enhance prompt effectiveness with more category-related information.
3) Extensive experiments on three commonly used evaluation settings, including base-to-new generalization, cross-dataset evaluation and domain generalization, demonstrate remarkable improvements with our method, better than existing state-of-the-art methods.

Despite achieving state-of-the-art performance, several challenging issues with HPT remain unresolved. First, HPT employs LLMs to generate category-related descriptions via handcrafted prompt templates. However, this approach may be ineffective as it does not guarantee that descriptions will be sufficiently discriminative among categories. Additionally, HPT models structured knowledge as matrices and integrates them additively into attention computation, which is suboptimal since it treats all relationships of the same type equally. Furthermore, despite its effectiveness in modeling linguistic knowledge, hierarchical prompt learning is prone to overfitting and could perform poorly when conducting generalization.

To further enhance our model's performance beyond that presented in our conference paper version~\citep{wang2024learning}, we propose HPT++ in Section 4. 
Specifically, we refine the knowledge generation process, producing and merging coarse-grained and fine-grained descriptions into multi-granularity descriptions for generating structured graphs with more discriminative semantics. 
Additionally, we experiment with various methods to model structured information and re-design the relationship-driven attention re-weighting module, enabling re-weighting of attention maps according to relationships between key elements with a predefined ratio. 
Finally, to avoid over-fitting in downstream generalization tasks, we incorporate a consistency constraint between prompted and pre-trained text encoders to learn more robust representations.
These improvements and comparisons to HPT are validated with extensive experiments.

\section{Related Work}

\subsection{Large Language Models}
Large Language Models (LLMs)~\citep{brown2020language, zhang2022opt, chowdhery2022palm, llama3modelcard, openai2023gpt4}, trained on extensive web-scale datasets, has gained widespread popularity due to its ability to generate text resembling human writing and to discern intricate patterns across diverse domains. Leveraging the vast potential of LLMs, recent studies have demonstrated their effectiveness in addressing various vision-language tasks~\citep{chen2022visualgpt, alayrac2022flamingo, cheng2023hybrid, yang2022zero}. Additionally, other studies investigate prompting vision-language models with LLMs for image classification, continuous learning, image caption generation, and action understanding~\citep{zhang2023prompt, li2022bridge, wang2022learning}. In this study, we aim to leverage the capabilities of LLMs in the field of image classification. When prompted with a target category, LLMs can generate related descriptions and corresponding structured relationships.

\subsection{Visual-Language Models}
Visual-language models (VLMs) have played a crucial role in advancing open vocabulary image classification, with CLIP~\citep{radford2021learning} pioneering this domain. Notable approaches include scaling up models with larger amounts of data, batch sizes, and model sizes, such as Align~\citep{jia2021scaling} and Basic~\citep{pham2021combined}; refining objective functions with models like SLIP~\citep{mu2022slip}, FILIP~\citep{yao2021filip}, and Lion~\citep{chen2023symbolic}; and incorporating supplementary information during training using models such as Florence~\citep{yuan2021florence}, UniCL~\citep{yang2022unified}, K-LITE~\citep{shen2022k}, and REACT~\citep{liu2023learning}. Our study is motivated by the goal of enhancing CLIP's capabilities through improved multi-modal prompts.

\subsection{Prompt Learning for Vision-Language Models}
Prompt learning originates in natural language processing (NLP) and aims to enhance interaction with large language models~\citep{liu2023pre, brown2020language, wei2022chain}. Some efforts~\citep{menon2022visual, pratt2022does} propose leveraging pre-trained linguistic knowledge from LLMs to generate prompts, thereby enhancing vision-language models without requiring additional training or labeling. To automate prompt engineering and explore optimal prompts, other studies~\citep{rao2022denseclip, zhou2022learning, zhou2022conditional, lu2022prompt} employ learnable text inputs, optimizing them during training, a process known as prompt tuning. With the emergence of visual prompt tuning (VPT)~\citep{jia2022visual}, recent methods~\citep{khattak2023maple, zhao2024learning} take a multi-modal approach, applying prompting to both modalities to improve alignment between vision and language representations. In contrast to prior studies, we generate diverse forms of linguistic knowledge and perform hierarchical prompt tuning based on this knowledge to produce more robust representations.

\section{HPT}
\subsection{Overall Pipeline}
We present the overall pipeline of our framework. 
As a baseline network, we apply a dual-path asymmetric network~\citep{zhao2024learning} for prompt tuning with visual-language models.
This network experts in addressing over-fitting issues of the learned prompts, particularly in few-shot learning scenarios. 
To perform prompt tuning for transformer-like encoders, learnable vectors are introduced at each Transformer layer’s input space as prompts. 
The framework incorporates a novel asymmetric contrastive loss, training the prompted image encoder and text encoder separately, using the frozen encoder from the counterpart modality as guidance.
Specifically, representations of prompted and frozen encoders from different modalities are aligned asymmetrically, generating two probabilities from the two frozen-prompted pairs, which are averaged to derive an overall probability.
All three probabilities are used to calculate the asymmetric loss $\mathcal{L}_{asy}$ for training, whereas only the overall probability is utilized during inference, following the previous work~\citep{zhao2024learning}.

For a specific category, we initially input a set of handcrafted templates filled with the category name as instruction into LLMs to generate human-like descriptions. 
Additionally, we input the generated descriptions, along with another instruction, into LLMs to capture the well-organized structure of each description, which includes category-related elements such as entities, attributes, and their relationships.
Detailed exposition is provided in Section \textbf{Linguistic Data Generation}.
Instead of modifying visual prompts, we focus primarily on prompt tuning for the text modality. 
Unlike the vanilla-prompted text encoder in the previous dual-path asymmetric network, Section \textbf{Hierarchical Prompt Tuning} offers a novel and detailed exploration of the core structure of this encoder for tuning prompts across different semantic levels. 
In particular, unstructured descriptions are fed into the frozen encoder, while relationship-guided graphs along with the corresponding category name are fed into the novel hierarchical prompted encoder, which is specifically designed and finetuned for modeling structured information. 
To effectively capture LLM-generated element-wise correspondences, the hierarchical prompted text encoder integrates a relationship-guided attention module, whose detailed implementation will be elaborated in Section \textbf{Relationship-guided Attention Module}.
\subsection{Linguistic Data Generation}
\begin{table*}[ht]
\centering
\caption{[CLASS] token and [TYPE] token for 11 image classification datasets. [X] denotes the category name. \protect\label{tab:a2}}
\resizebox{12.5cm}{!}{
\begin{tabular}{l|l|l}
    \toprule  
     \textbf{Dataset} & \textbf{[CLASS]} & \textbf{[TYPE]}\\
    \midrule 
    ImageNet & [X] & objects \\
    OxfordPets & a pet [X] & types of pets \\ 
    Caltech101 & [X] & objects \\
    DescribableTextures & a [X] texture & types of texture \\
    EuroSAT & [X] & types of land in a centered satellite photo \\
    FGVCAircraft & a [X] aircraft & types of aircraft \\
    Food101 & [X] & types of food \\
    OxfordFlowers & a flower [X] & types of flowers \\
    StanfordCars & a [X] car & types of car \\
    SUN397 & a [X] scene & types of scenes \\
    UCF101 & a person doing [X] & types of action \\
    \bottomrule 
\end{tabular}}
\end{table*}
To acquire linguistic knowledge, we use one of the most powerful LLMs, ChatGPT~\citep{openai2023gpt4}, to generate descriptions with corresponding structured relationships. As shown in Figure~\ref{fig:intro}, we adopt $N_h$ question templates as the language instruction $T$ for LLMs, e.g., ``What does a [CLASS] look like among all a [TYPE]?" or ``What are the distinct features of [CLASS] for recognition among all [TYPE]?", etc. [CLASS] denotes a specific category name with a modifier, like ``a pet Abyssinian". [TYPE] indicates the type of objects related to the dataset, like ``types of pets" for OxfordPets~\citep{parkhi2012cats}. A full list of [CLASS] token and [TYPE] token for all datasets is illustrated in Table~\ref{tab:a2}. We denote the generated descriptions from $T$ as $D=\{d_i\}_{i=1}^{N_h}$, formulated as:
\begin{equation}
D = \operatorname{LLM}(T).
\end{equation}

For descriptions in $D$, we design an extra instruction $T^{'}$ to leverage LLMs for producing structured knowledge, including entities, attributes, and relationships among them. We denote the structured knowledge generated from $D$ as $R$, expressed as:
\begin{equation}
R = \operatorname{LLM}([T^{'}, D]).
\end{equation}
\noindent Here $R=\{r_i\}_{i=1}^{N_h}$, $r_i=\{E_i, A_i, R_{e2e, i}, R_{e2a, i}\}$, where $E_i$, $A_i$, $R_{e2e, i}$, $R_{e2a, i}$ represent the entity set, the attribute set, the set of entity-entity relationships, and the set of entity-attribute relationships based on description $d_i$.

Our method utilizes both descriptions $D$ and structured knowledge $R$ as the source of category-related textual information, leading to effective prompt tuning.

\begin{figure*}[t]  
\begin{center}  
\subfigure[Overall pipeline for hierarchical prompt tuning]{  
\includegraphics[width=0.44\linewidth]{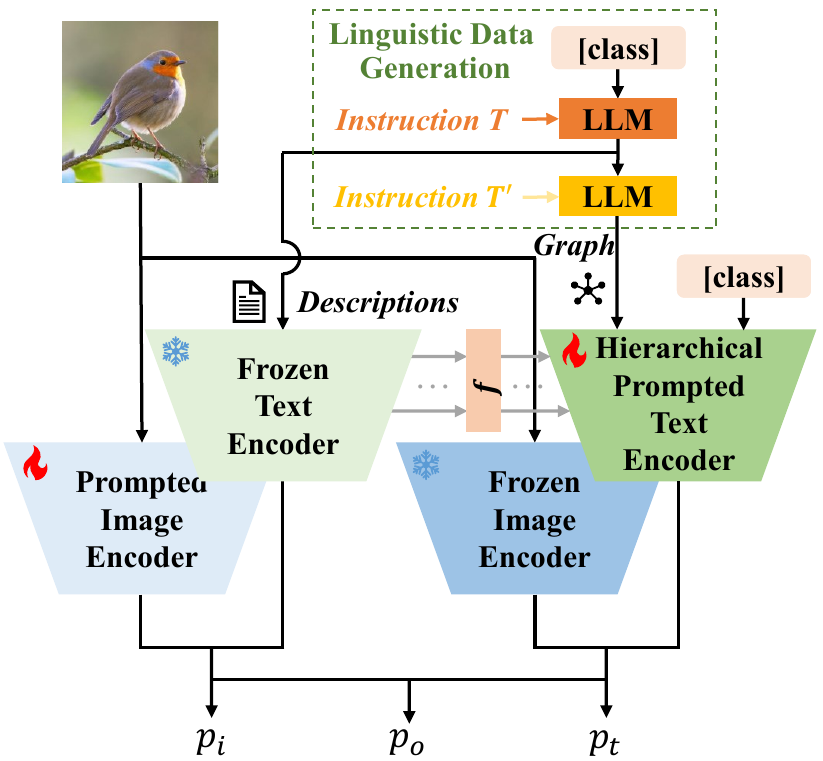}}\quad\quad
\subfigure[Structure of hierarchical prompted text encoder]{  
\includegraphics[width=0.47\linewidth]{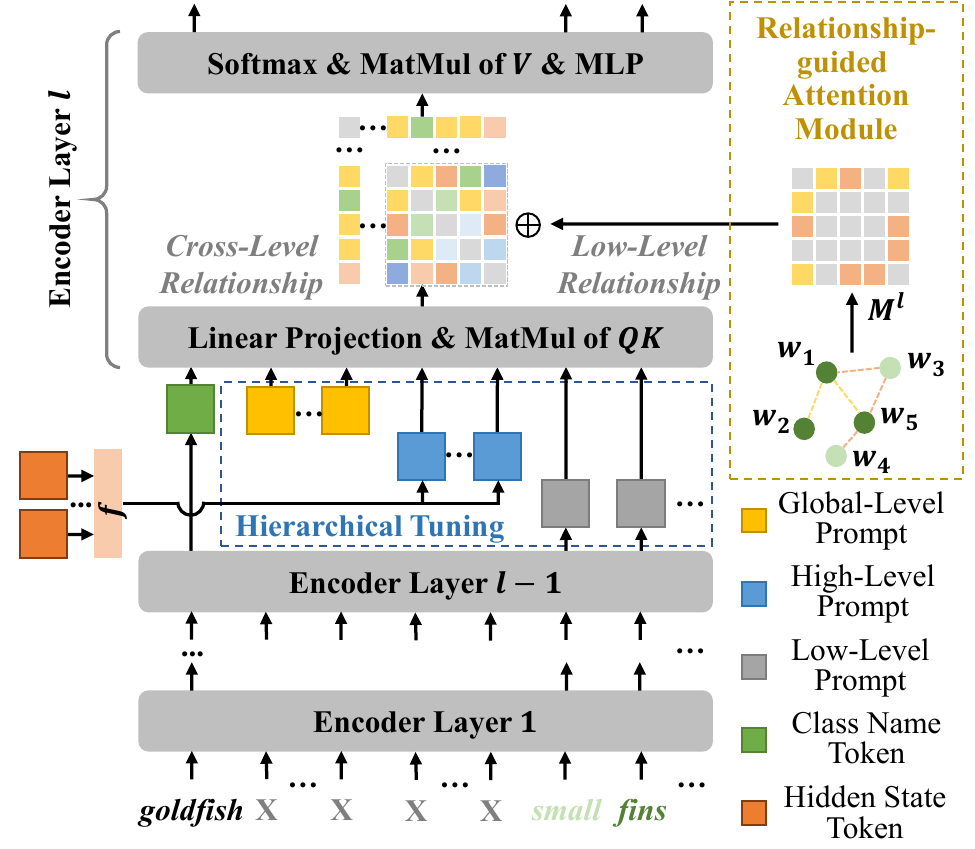}}
\end{center}  
\caption{Our HPT applies a dual-path asymmetric network as the framework. Descriptions and relationship-guided graphs with class names are used as input for the frozen text encoder and the hierarchical prompted text encoder respectively. In the hierarchical prompted text encoder, we apply three types of prompts, low-level prompts, high-level prompts, and global-level prompts for hierarchical tuning, and design a relationship-guided attention module for modeling structured knowledge. \protect\label{fig:framework}}  
\end{figure*}

\subsection{Hierarchical Prompt Tuning}
Given descriptions $D$ and structured knowledge $R$, we aspire to simultaneously model both structured and conventional linguistic knowledge. Therefore, we propose a novel approach called Hierarchical Prompt Tuning (HPT), which leverages both forms of knowledge for learning prompts in a hierarchical manner, as shown in Figure \ref{fig:framework}(b). HPT contains low-level prompts, high-level prompts, and global-level prompts, respectively denoted as $p_l$, $p_h$, $p_g$. 

\noindent\textit{Low-Level Prompt} To model pair-wise relationships within a description, we select essential words from this description as the input of the text encoder. Specifically, for entities in the entity set $E_i$ and attributes in the attribute set $A_i$, we simply concatenate them together as the low-level prompts $p_l^0$ for description $d_i$ and feed them into the first layer of the encoder. These prompts are seen as nodes in a relationship-guided graph, whose relationships are further processed by a novel relationship-guided attention module.

\noindent\textit{High-Level Prompt} In order to capture more intricate associations between individual tokens and the complete description, we derive high-level prompts $p_h$ that encapsulate the overall semantics of the category based on a series of descriptions. In detail, we feed descriptions $D$ into the frozen text encoder. Instead of simply utilizing representations from the last layer, we extract the last tokens from each layer containing rich semantics and feed them into a learnable prompt generator $f$, represented as:
\begin{equation}
p_{h,i}^l = f\left(h_i^l\right), 
\end{equation}
where $h_i^l$ represents the last token of description $d_i$ at the $l$-th layer. These tokens are then concatenated together as the high-level prompts $p_h^l=[p_{h,1}^l; ...; p_{h,{N_h}}^l ]$ of this category, which are further integrated into the corresponding layer of the hierarchical prompted encoder.

\noindent\textit{Global-Level Prompt} To represent category-shared knowledge pertinent to the task, we employ the standard approach for tuning the global-level prompts $p_g$. Instead of leveraging any form of knowledge, we automatically learn $N_g$ category-agnostic continuous vectors shared across categories as contexts and concatenate them with other prompts for each layer. 

\noindent\textit{Hierarchical Tuning} Based on the above prompts, we conduct the proposed hierarchical prompt tuning on the hierarchical prompted text encoder, formulated as
\begin{align} {\left[c^1, \_, \_, p_{l}^1\right] } & =L_1\left(\left[c, p_{g}^0, p_{h}^0, p_{l}^0\right]\right)\nonumber \\ {\left[c^{i}, \_, \_, p_{l}^{i}\right]} & =L_i\left(\left[c^{i-1}, p_{g}^{i-1}, p_{h}^{i-1}, p_{l}^{i-1}\right]\right), \\ i & = 2, 3, ..., N\nonumber\end{align}
where $c$ represents the token of the class name. Via the projection head of the text encoder $\operatorname{TextProj}$, the final text representation $z$ is acquired by projecting the text embeddings $x^N$ corresponding to the last token of the last transformer block $L_N$ to a common V-L latent embedding
space:
\begin{equation}
z = \operatorname{TextProj} \left( x^N\right).
\end{equation}
\subsection{Relationship-guided Attention Module}
We introduce a relationship-guided attention module to model structured knowledge $R$ to capture pair-wise correspondences among entities and attributes in a layer-wise manner. 
For the $l$-th layer of a transformer-like encoder, an attention-based matrix $M^l$ is constructed based on generated relationships from each description. 
Two types of scalar values $\lambda_{e2e}^l$ and $\lambda_{e2a}^l$ are learned to indicate the strength of the relationship of entity-entity pairs and entity-attribute pairs separately. We assign the value to the respective element in the matrix, formulated as:
\begin{equation}
M_{i,j}^l=\left\{
\begin{array}{ll}
\lambda_{e2e}^l & \ \ \ {(w_i, w_j) \in R_{e2e}}\\
\lambda_{e2a}^l & \ \ \ {(w_i, w_j) \in R_{e2a}}\\
0 & \ \ \ \mathrm{otherwise,}
\end{array}
\right.
\end{equation}
where $w_i$ indicates the entity or attribute associated with the $i$-th token in the sequence of low-level prompts.

Guided by structured knowledge, the learned attention-based matrices are integrated into layers of the text encoder. 
In practice, we compute the attention function on a set of queries simultaneously, packed together into a matrix $Q$. The keys and values are also packed together into matrices $K$ and $V$. For the $l$-th layer, with the attention-based matrix $M^l$, the output of self-attention is computed as:
\begin{equation}
\operatorname{Attention}^l(Q, K, V)=\operatorname{softmax}\left(\frac{Q K^\top+M^l}{\sqrt{d_k}}\right) V.
\label{eq:8}
\end{equation}
By explicitly adding $M^l$ into the calculation of self-attention, our model explicitly represents rich structured relationships within each description, thus enhancing crucial information associated with the category.

To deal with more intricate relationships, we include high-level and global-level prompts for the construction of long-term relationships. Unlike modeling correspondences with matrices, we automatically leverage the implicit associations through cross-level self-attention itself without any manual intervention. This design, as a hierarchical knowledge modeling approach, blends holistic semantics from multiple levels with structured relationships, thereby helping us discover complex associations that LLMs have failed to identify.

\section{HPT++}
\subsection{Overall Improvements}
HPT, as introduced, can simultaneously model both structured and conventional linguistic knowledge. 
This capability makes it effective for handling complex and long-term relationships. 
We next explore several improvements to the original framework while maintaining the hierarchical structure. 
Specifically, in Section \textbf{Multi-Granularity Knowledge Generation}, we refine the knowledge generation process by merging coarse-grained and fine-grained descriptions into multi-granularity descriptions to create structured graphs with richer semantics. 
Additionally, we experiment with various strategies to model structured information and redesign the relationship-driven attention re-weighting module, allowing attention intensity to be scaled based on the generated relationships at a preset ratio. Detailed exposition is provided in Section \textbf{Relationship-Driven Attention Re-Weighting Module}. 
Finally, to avoid overfitting in downstream generalization tasks, we incorporate a consistency constraint between hierarchical prompted and pre-trained text encoders to learn robust and adaptive representations, detailed in Section \textbf{Consistency Constraint on Hierarchical Prompted Text Encoder}.
\begin{figure}[t]
    \centering
    \center{\includegraphics[width=7.6cm]{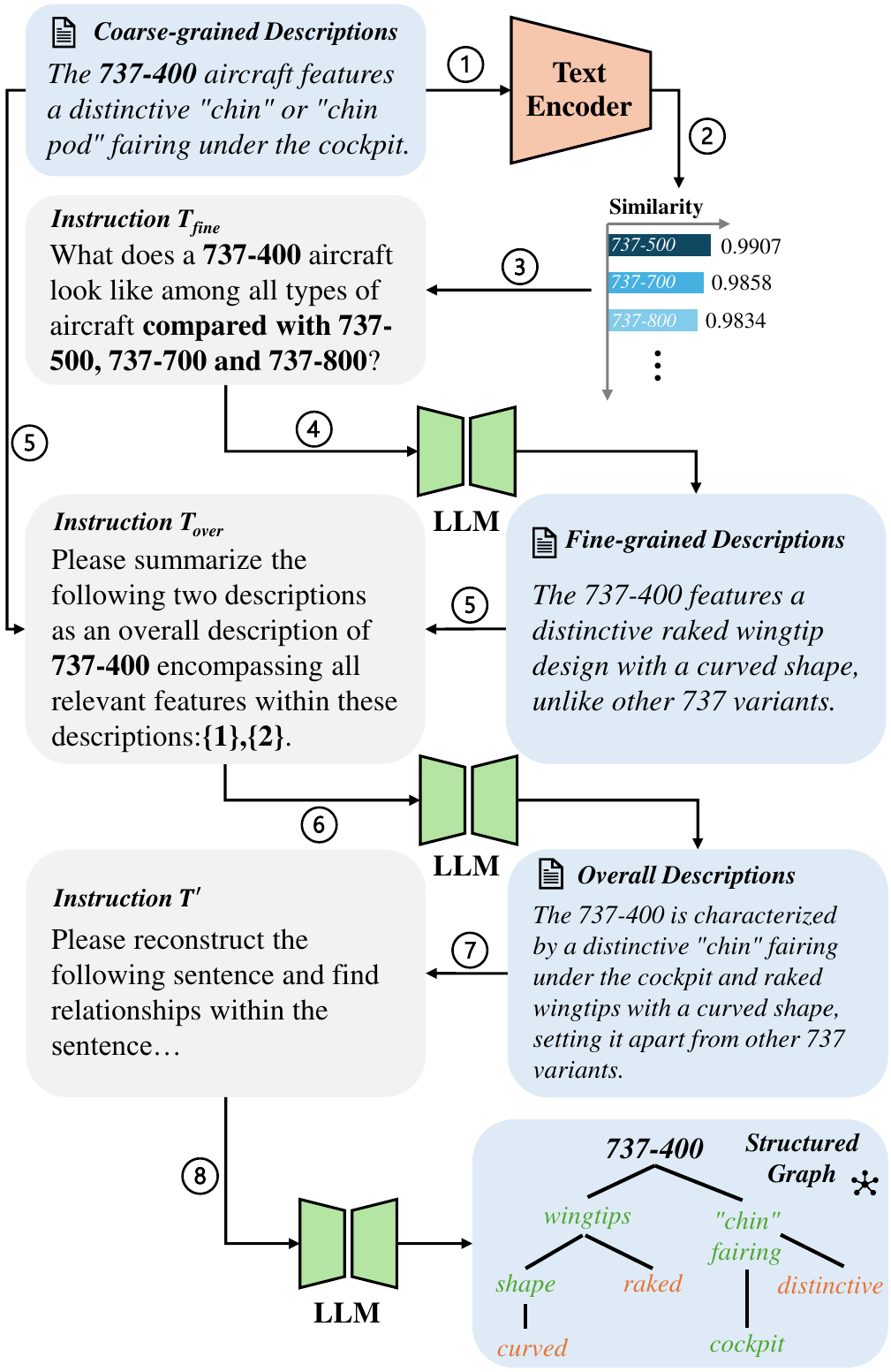}}
    \caption{Illustration of multi-granularity knowledge generation. We firstly compute the similarity between coarse-grained descriptions of different categories, and then generate fine-grained descriptions for each category based on its closest categories. We integrate descriptions of both granularities to produce an overall description with multi-granularity semantics, which is subsequently used for generating structured graphs. \protect\label{fig:mgc}}
\end{figure} 
\subsection{Multi-Granularity Knowledge Generation}
HPT leverages LLMs to generate category-related descriptions based on handcraft prompt templates. This approach seems suboptimal since it does not ensure that the generated descriptions will be sufficiently discriminative to distinguish between different categories. 
For instance, in FGVCAircraft, the descriptions of categories such as ``737-400", ``737-500", ``737-700", and ``737-800" all share the characteristic of ``a distinctive `chin' fairing under the cockpit," as these categories represent different variants of Boeing 737 series. This similarity makes it challenging for VLMs to correctly identify the category of an image.
To prevent highly similar descriptions among certain categories, we propose multi-granularity knowledge generation, as shown in Figure~\ref{fig:mgc}. 

We firstly compute the similarity between descriptions of different categories using the pre-generated descriptions introduced in HPT (referred to as coarse-grained descriptions) and then generate fine-grained descriptions for each category based on its closest categories. 
Given \(D^c_{coar}=\{d^c_{coar, i}\}_{i=1}^{N_h}\) as the coarse-grained descriptions of the \(c\)-th category, we input \(d^c_{coar, i}\) into the frozen text encoder to obtain the corresponding text representation \(h^c_i\), which is then normalized to \(\tilde{h}^c_i\). We average the normalized representations for the \(c\)-th category as $\overline{h}^c=\frac{1}{N_h}\sum^{N_h}_{i=1}\tilde{h}^c_i$. We use cosine distance to find the top-\(C\) relevant classes for each category, denoted as [CLASS]\(_0\), ..., [CLASS]\(_{C-1}\). We reuse \(N_h\) question templates as $T_{fine}$, and for each handcrafted question template, we append ``compared with [CLASS]\(_0\), ..., [CLASS]\(_{C-1}\)" as $T_{fine}(c)$ to prompt LLMs to generate fine-grained descriptions $D^c_{fine} =\{d^c_{fine, i}\}_{i=1}^{N_h}$ for the $c$-th category distinctive to other similar categories, expressed as:
\begin{equation}
D^c_{fine} = \operatorname{LLM}(T_{fine}(c)).
\end{equation}

Furthermore, we integrate descriptions of both granularities to produce a comprehensive description with multi-granularity semantics, which is subsequently used for generating structured knowledge. Specifically, we design an instruction like ``Please summarize the following two descriptions as an overall description of [CLASS] encompassing all relevant features within these descriptions: \{$d_1$\}, \{$d_2$\}.'' as $T_{over}(d_1, d_2)$ for LLMs to output a summarized description. The process of obtaining the overall description corpus for the $c$-th category is represented as:
\begin{equation}
D^c_{over} = \operatorname{LLM}(T_{over}(D^c_{coar}, D^c_{fine})).
\end{equation}

For descriptions in $D^c_{over}$, we revise the instruction $T^{'}$ in HPT to leverage LLMs for producing structured knowledge in a simpler form, only including relationships. We denote the relationships generated from $D^c_{over}$ as $R^c_{over}$, formulated as:
\begin{equation}
R^c_{over} = \operatorname{LLM}([T^{'}, D^c_{over}]).
\end{equation}
Here $R^c_{over}=\{r^c_i\}_{i=1}^{N_h}$, where $r^c_i$ represent the set of relationships based on the description $d^c_{over, i}$, including the subject, the verb and the direct object (or attribute) for each relationship.
HPT++ utilizes both descriptions $D_{over}$ and structured knowledge $R_{over}$ as the final corpus to provide multi-granularity textual information for each category, leading to effective prompt tuning.
It should be noted that unlike HPT, HPT++ does not generate entities and attributes for each description. Instead, the description itself is directly input into the hierarchical prompted text encoder, identical to the input of the frozen text encoder. The tokens in the description serve as the low-level prompt and are further processed by our proposed relationship-driven attention re-weighting module, using the generated structured graph as connections.

\subsection{Relationship-Driven Attention Re-Weighting Module}
In HPT, relationships are modeled as matrices and applied additively to attention computation based on the proposed relationship-guided attention module. We believe this approach is suboptimal, as the intensity of relationships within the same type is added equally (i.e., by adding the same scalar to the attention map). To address this issue, we investigate different approaches to modeling relationships and propose a relationship-driven attention re-weighting module. Our findings indicate that re-weighting with element-wise multiplication enhances recognition performance more effectively than simple addition, thereby better integrating structured knowledge into our model.

During the construction of the re-weighting matrix $M^l$ for the $l$-th layer of the encoder, HPT++ uses a preset hyperparameter $\beta$ to indicate the re-weighting intensity, instead of learning two scalar values $\lambda_{e2e}^l$ and $\lambda_{e2a}^l$ in HPT. When $\beta = 0$, no re-weighting operations are performed. As $\beta$ increases, the attention intensity between tokens within a relationship is amplified, while that between tokens without a relationship diminishes.
Based on the given relationship set $R$, we assign the value to the respective element in the matrix, expressed as:
\begin{equation}
M_{i,j}^l=\left\{
\begin{array}{ll}
1 + \beta & \ \ \ (w_i, w_j) \in R\\
\frac{1}{1 + \beta} & \ \ \ \mathrm{otherwise,}
\end{array}
\right.
\end{equation}
where $w_i$ and $w_j$ indicates the $i$-th and the $j$-th token in the sequence of the corresponding description. This re-weighting method proportionally scales the attention intensity between tokens, thereby preserving the relative intensity in the original attention map. We re-formulate Equation \ref{eq:8} as follows:
\begin{equation}
\operatorname{Attention}^l(Q, K, V)=\operatorname{softmax}\left(\frac{Q K^\top \odot M^l}{\sqrt{d_k}}\right) V.
\label{eq:13}
\end{equation}

We also investigate and experiment with alternative re-weighting schemes for comparison, such as enhancing only the elements related to a relationship while keeping the other elements unchanged. Detailed studies can be found in the experimental section.
\subsection{Consistent Constraint on Hierarchical Prompted Text Encoder}
Although hierarchical prompt learning effectively models linguistic knowledge, it still faces potential overfitting issues and exhibits room for improvement in generalizing to new categories or domains. Inspired by PromptSRC~\citep{khattak2023self} and CoPrompt~\citep{roy2023consistency}, we impose a consistency constraint on the text branch, using cosine similarity between representations from the pre-trained and hierarchical prompted text encoders to regularize our hierarchical prompts. The asymmetric network mentioned earlier aligns the hierarchical text branch with the pre-trained visual branch, while the consistency constraint aligns it with the pre-trained text branch. This dual alignment enhances the robustness and generalization of the learned representations by leveraging pre-trained knowledge from both modalities. Furthermore, to enhance learning capacity and improve adaptation, we introduce an adapter $\phi$ at the top of the hierarchical prompted text encoder. This adapter consists of trainable parameters designed to transform the embedding vector~\citep{gao2024clip}.
The consistency loss is represented as:
\begin{equation}
\mathcal{L}_c = 1 - \frac{\phi(z) \cdot \Theta(t)}{\|\phi(z)\| \|\Theta(t)\|}.
\end{equation}
Here $\Theta$ denotes the pre-trained text encoder and $t$ stands for the input description. The final training loss $\mathcal{L}$ of HPT++ is obtained by summing the asymmetric loss and the consistency loss with a balancing ratio $\lambda$, which is formulated as:
\begin{equation}
\mathcal{L} = \mathcal{L}_{asy} + \lambda\mathcal{L}_{c}.
\end{equation}

\section{Experimental Setup}
To evaluate our method, we follow the experiment setup established in previous works~\citep{zhou2022learning, zhou2022conditional}. We first describe evaluation protocols and datasets, followed by a discussion on implementation details.

\subsection{Evaluation Protocols}
\noindent\textit{Base-to-New Generalization} Aiming to evaluate the generalizability across various classes, this process involves dividing the dataset into base (seen) and new (unseen) classes and then training the model using a small number of samples from the base classes. Finally, we evaluate the model's performance on both base (few-shot performance) and new (zero-shot performance) classes. Additionally, we calculate the harmonic mean over the accuracy on both base and new classes to highlight the generalization trade-off.

\noindent\textit{Cross-Dataset Evaluation} This evaluation approach aims to assess the zero-shot ability of the model on a cross-dataset setup. To validate the potential of our approach in cross-dataset transfer, we train our model on all ImageNet classes in a few-shot manner and evaluate it directly on ten other unseen datasets with unknown categories in a zero-shot regime. 

\noindent\textit{Domain Generalization} To evaluate the robustness of our method on out-of-distribution datasets, we consider ImageNet as the source domain and its other variants as the target domain. We finetune our model on ImageNet in a few-shot setting and evaluate it on four variants of ImageNet with identical classes or subsets while manifesting diverse domain shifts.

\subsection{Datasets}
For base-to-new generalization and cross-dataset evaluation, we evaluate the performance of our method on 11 image recognition datasets, which cover a wide range of recognition tasks. Specifically, the benchmark includes ImageNet~\citep{deng2009imagenet} and Caltech101~\citep{fei2004learning} for classification on generic objects; OxfordPets~\citep{parkhi2012cats}, StanfordCars~\citep{krause20133d}, Flowers102~\citep{nilsback2008automated}, Food101~\citep{bossard2014food} and FGVCAircraft~\citep{maji2013fine} for fine-grained classification; SUN397~\citep{xiao2010sun} for scene recognition; UCF101~\citep{soomro2012ucf101} for action recognition; DTD~\citep{cimpoi2014describing} for texture classification; and finally EuroSAT~\citep{helber2019eurosat} for satellite imagery recognition. For domain generalization, we utilize ImageNet as the source dataset and its four variants as target datasets including ImageNetV2~\citep{recht2019imagenet}, ImageNet-Sketch~\citep{wang2019learning}, ImageNet-A~\citep{hendrycks2021natural} and ImageNet-R~\citep{hendrycks2021many}.

\subsection{Implementation Details}
We apply prompt tuning to the pre-trained CLIP model~\citep{radford2021learning}, using ViT-B/16 as the visual backbone. We employ SGD optimization with an initial learning rate of 0.0025 for base-to-new generalization and 0.001 for other tasks, using a batch size of 8. The maximum number of epochs is set to 10 for base-to-new generalization. For other tasks, we train our model for 3 epochs for HPT and 5 epochs for HPT++. The length of global-level prompts $N_g$ is set to 2 at each layer, and the number of descriptions per category $N_h$, which also corresponds to the length of high-level prompts, is set to 5. We randomly select one description per category to conduct relationship modeling at each step during training to optimize memory usage, while leveraging all $N_h$ descriptions per category for inference. We use GPT-3.5-turbo~\citep{openai2023gpt4} and Llama-3-8B~\citep{llama3modelcard} as LLMs in our study, both of which have comparable performance. Our research on the performance of different LLMs will be presented in the experimental section.

For HPT++, assuming $N_c$ is the number of categories in a dataset, we determine the number of closest categories $C$ for fine-grained description generation using the following function:
\begin{equation}
C = \lfloor lg(N_c)\rfloor+1.
\end{equation}
The re-weighting intensity ratio $\beta$ is set to 0.2, and the balancing ratio $\lambda$ for the training loss is set to 1. Following prior works, we select 16 shots for training and use the entire test set for evaluation. For domain generalization and cross-dataset evaluation, we use the same hyperparameters across datasets, avoiding a separate search in CoPrompt~\citep{roy2023consistency}.

\section{Experimental results}
We evaluate our approach in three generalization settings, i.e. base-to-new generalization, cross-dataset evaluation, and domain generalization. We compare its performance with zero-shot CLIP~\citep{radford2021learning} and recent prompt learning works as strong baselines~\citep{zhou2022learning,zhou2022conditional,zhu2023prompt,yao2023visual,khattak2023self,zhao2024learning,roy2023consistency}. In the case of CLIP, we use handcrafted prompts specifically designed for each dataset. We further conduct several ablation experiments and sample analyses to better demonstrate the effectiveness of our proposed hierarchical prompt tuning.
\begin{sidewaystable*}
\small
\centering
\caption{Comparison with existing methods on base-to-new generalization. B: Base Classes. N: New Classes. HM: Harmonic mean. HPT and HPT++ demonstrate strong generalization performance on 11 image recognition datasets. \protect\label{tab:b2n}}
\setlength{\tabcolsep}{0.8mm}{\begin{tabular*}{\textheight}{@{\extracolsep\fill}lccccccccccccc}
\toprule 
Method & & ImageNet & Caltech & Pets & Cars & Flowers & Food101 & Aircraft & SUN397 & DTD & EuroSAT & UCF101 & \textit{Average} \\
\midrule 
 & B & 72.43 &	96.84 &	91.17 &	63.37 &	72.08 &	90.10 &	27.19 &	69.36 &	53.24 &	56.48 &	70.53 &	69.34 \\
 CLIP~\citep{radford2021learning} & N & 68.14 &	94.00 &	97.26 &	74.89 &	77.80 &	91.22 &	36.29 &	75.35 &	59.90 &	64.05 &	77.50 &	74.22 \\
 & H & 70.22 &	95.40 &	94.12 &	68.65 &	74.83 &	90.66 &	31.09 &	72.23 &	56.37 &	60.03 &	73.85 &	71.70 \\
\midrule 
 & B & 76.47 &	98.00 &	93.67 &	78.12 &	97.60 &	88.33 &	40.44 &	80.60 &	79.44 &	92.19 &	84.69 &	82.69 \\
CoOp~\citep{zhou2022learning} & N & 67.88 &	89.81 &	95.29 &	60.40 &	59.67 &	82.26 &	22.30 &	65.89 &	41.18 &	54.74 &	56.05 &	63.22 \\
 & H & 71.92 &	93.73 &	94.47 &	68.13 &	74.06 &	85.19 &	28.75 &	72.51 &	54.24 &	68.69 &	67.46 &	71.66 \\
\midrule 
 & B & 75.98 &	97.96 &	95.20 &	70.49 &	94.87 &	90.70 &	33.41 &	79.74 &	77.01 &	87.49 &	82.33 &	80.47 \\
CoCoOp~\citep{zhou2022conditional} & N & 70.43 &	93.81 &	97.69 &	73.59 &	71.75 &	91.29 &	23.71 &	76.86 &	56.00 &	60.04 &	73.45 &	71.69 \\
 & H & 73.10 &	95.84 &	96.43 &	72.01 &	81.71 &	90.99 &	27.74 &	78.27 &	64.85 &	71.21 &	77.64 &	75.83 \\
\midrule 
 & B & 77.02 &	98.02 &	95.07 &	77.68 &	95.54 &	90.37 &	40.54 &	81.26 &	77.35 &	90.11 &	84.33 &	82.48 \\
ProGrad~\citep{zhu2023prompt} & N & 66.66 &	93.89 &	97.63 &	68.63 &	71.87 &	89.59 &	27.57 &	74.17 &	52.35 &	60.89 &	74.94 &	70.74 \\
 & H & 71.46 &	95.91 &	96.33 &	72.88 &	82.03 &	89.98 &	32.82 &	77.55 &	62.45 &	72.67 &	79.35 &	76.16 \\
\midrule 
 & B & 75.83 &	97.72 &	94.65 &	71.76 &	95.00 &	90.50 &	36.21 &	80.29 &	77.55 &	85.64 &	82.89 &	80.73 \\
KgCoOp~\citep{yao2023visual} & N & 69.96 &	94.39 &	97.76 &	\textbf{75.04} &	74.73 &	91.70 &	33.55 &	76.53 &	54.99 &	64.34 &	76.67 &	73.61 \\
 & H & 72.78 &	96.03 &	96.18 &	73.36 &	83.65 &	91.09 &	34.83 &	78.36 &	64.35 &	73.48 &	79.65 &	77.00 \\
\midrule 
 & B & 76.66 &	97.74 &	95.43 &	72.94 &	95.92 &	90.71 &	37.44 &	80.82 &	80.36 &	94.07 &	83.00 &	82.28 \\
MaPLe~\citep{khattak2023maple} & N & 70.54 &	94.36 &	97.76 &	74.00 &	72.46 &	92.05 &	35.61 &	78.70 &	59.18 &	73.23 &	78.66 &	75.14 \\
 & H & 73.47 &	96.02 &	96.58 &	73.47 &	82.56 &	91.38 &	36.50 &	79.75 &	68.16 &	82.35 &	80.77 &	78.55 \\
\midrule 
 & B & 77.39 &	98.28 &	95.71 &	75.43 &	97.53 &	\textbf{90.76} &	39.38 &	82.10 &	82.52 &	93.37 &	84.70 &	83.38 \\
PromptSRC~\citep{khattak2023self} & N & 71.06 &	94.58 &	96.98 &	74.43 &	74.54 &	91.77 &	37.59 &	79.01 &	60.10 &	78.34 &	78.56 &	76.09 \\
 & H & 74.09 &	96.39 &	96.34 &	74.93 &	84.50 &	91.26 &	38.46 &	80.53 &	69.55 &	85.20 &	81.51 &	79.57 \\
 \midrule 
 & B & 77.60 &	\textbf{98.10} &	95.33 &	\textbf{78.27} &	98.07 &	90.67 &	\textbf{42.73} &	\textbf{82.67} & 83.37 & 92.90 & \textbf{87.10} & 84.26 \\
MetaPrompt~\citep{zhao2024learning} & N & 70.73 &	94.03 &	97.30 &	74.97 &	76.50 &	91.53 &	37.87 &	78.47 &	62.97 &	73.90 &	78.80 &	76.10 \\
 & H & 74.01 &	96.02 &	96.30 &	\textbf{76.58} &	85.95 &	91.10 &	40.15 &	80.52 &	71.75 &	82.32 &	82.74 &	79.97 \\
\midrule 
 & B         & 77.67 & 98.27 & 95.67 & 76.97 & 97.27 & 90.73 & 40.20 & 82.63 & 83.13 & 94.60 & 86.90 & 84.00 \\
CoPrompt~\citep{roy2023consistency} & N & \textbf{71.27} & 94.90 & 98.10 & 74.40 & 76.60 & \textbf{92.07} & 39.33 & \textbf{80.03} & 64.73 & 78.57 & 79.57 & 77.23 \\
 & H         & \textbf{74.33} & 96.55 & 96.87 & 75.66 & 85.71 & \textbf{91.40} & 39.76 & \textbf{81.31} & 72.79 & 85.84 & 83.07 & 80.48 \\
\midrule 
 & B             & \textbf{77.95} & \textbf{98.37} & 95.78 & 76.95 & \textbf{98.17} & 90.46 & 42.68 & 82.57 & 83.84 & 94.24 & 86.52 & \textbf{84.32}\\
\textbf{HPT}~\citep{wang2024learning} & N & 70.74 & 94.98 & 97.65 & 74.23 & \textbf{78.37} & 91.57 & 38.13 & 79.26 & 63.33 & 77.12 & 80.06 & 76.86\\
 & H             & 74.17 & 96.65 & 96.71 & 75.57 & \textbf{87.16} & 91.01 & 40.28 & 80.88 & 72.16 & 84.82 & 83.16 & 80.42\\
 \midrule 
 & B               & 77.66 & 98.17 & \textbf{95.94} & 76.99 & 97.50 & 90.56 & 40.50 & 82.40 & \textbf{84.18} & \textbf{95.31} & 86.26 & 84.13\\
\textbf{HPT++} & N & 71.11 & \textbf{95.78} & 97.89 & 74.24 & 76.69 & 91.62 & \textbf{42.19} & 79.86 & \textbf{66.39} & \textbf{80.64} & \textbf{81.50} & \textbf{77.99}\\
 & H               & 74.24 & \textbf{96.96} & \textbf{96.91} & 75.59 & 85.85 & 91.09 & \textbf{41.33} & 81.11 & \textbf{74.23} & \textbf{87.36} & \textbf{83.81} & \textbf{80.95}\\
\bottomrule
\end{tabular*}}
\end{sidewaystable*}

\begin{sidewaystable}
\small
\centering
\caption{Comparison with existing methods on cross-dataset evaluation. The best results are highlighted in bold while the second best results are marked with an underline. HPT and HPT++ achieve competitive performance providing the highest average accuracy, indicating superior generalization abilities on other datasets. \protect\label{tab:cde}}
\begin{tabular}{lcccccccccccc}
\toprule
& \multicolumn{1}{c}{\textbf{Source}} & \multicolumn{11}{c}{\textbf{Target}} \\
% \midrule
\cmidrule(lr){2-2} \cmidrule(lr){3-13}
& ImageNet & Caltech & Pets & Cars & Flowers & Food101 & Aircraft & SUN397 & DTD & EuroSAT & UCF101 & \textit{Average} \\
\midrule 
CoOp & 71.51 & 93.70 & 89.14 & 64.51 & 68.71 & 85.30 & 18.47 & 64.15 & 41.92 & 46.39 & 66.55 & 63.88 \\
CoCoOp & 71.02 & 94.43 & 90.14 & 65.32 & 71.88 & 86.06 & 22.94 & 67.36 & 45.73 & 45.37 & 68.21 & 65.74 \\
PromptSRC & 71.27 & 93.60 & 90.25 & 65.70 & 70.25 & 86.15 & 23.90 & 67.10 & 46.87 & 45.50 & 68.75 & 65.81 \\
MaPLe & 70.72 & 93.53 & 90.49 & 65.57 & 72.23 & 86.20 & 24.74 & 67.01 & 46.49 & 48.06 & 68.69 & 66.30 \\
CoPrompt & 70.80 & \textbf{94.50} & 90.73 & \underline{65.67} & 72.30 & \textbf{86.43} & 24.00 & 67.57 & 47.07 & \textbf{51.90} & 69.73 & 67.00 \\
\midrule 
\textbf{HPT} & \underline{71.72} & \underline{94.20} & \textbf{92.63} & \textbf{66.33} & \textbf{74.84} & 86.21 & \underline{25.68} & \underline{68.75} & 50.87 & 47.36 & \underline{70.50} & \underline{67.74} \\
\textbf{HPT++} & \textbf{71.81} & 94.02 & \underline{92.16} & 65.55 & \underline{72.43} & \underline{86.34} & \textbf{28.6} & \textbf{68.78} & \textbf{51.02} & \underline{50.76} & \textbf{70.53} & \textbf{68.02}\\
\bottomrule
\end{tabular}
\end{sidewaystable}

\begin{table}[!t]
\centering
\caption{Comparison with existing methods on domain generalization. The best results are highlighted in bold while the second best results are marked with an underline. Overall, HPT and HPT++ show consistent improvements on target variant datasets while achieving high accuracy on the source ImageNet dataset. \protect\label{tab:dg}}
\setlength{\tabcolsep}{0.7mm}{
\begin{tabular}{lcccccc}
\toprule
& \multicolumn{1}{c}{\textbf{Source}} & \multicolumn{5}{c}{\textbf{Target}} \\
\cmidrule(lr){2-2} \cmidrule(lr){3-7}
& ImNet & V2 & S & A & R & Avg.\\
\midrule 
CLIP & 66.73 & 60.83 & 46.15 & 47.77 & 73.96 & 57.17\\
CoOp & 71.51 & 64.20 & 47.99 & 49.71 & 75.21 & 59.28\\
CoCoOp & 71.02 & 64.07 & 48.75 & 50.63 & 76.18 & 59.90\\
MaPLe & 70.72 & 64.07 & 49.15 & 50.90 & 76.98 & 60.26\\
CoPrompt & 70.80 & 64.25 & \underline{49.43} & 50.50 & 77.51 & 60.42\\
PromptSRC & 71.27 & 64.35 & \textbf{49.55} & \underline{50.90} & \textbf{77.80} & 60.65\\
\midrule 
\textbf{HPT} & \underline{71.72} & \underline{65.25} & 49.36 & 50.85 & 77.38 & \underline{60.71}\\
\textbf{HPT++} & \textbf{71.81} & \textbf{65.31} & 49.28 & \textbf{51.18} & \underline{77.52} & \textbf{60.82}\\
\bottomrule
\end{tabular}}
\end{table}
\subsection{Base-to-New Generalization}
Table \ref{tab:b2n} presents the performance of various prompt tuning methods in base-to-new generalization setting on 11 recognition datasets.
Compared to the previous SOTA, CoPrompt, HPT demonstrates comparable performance on base classes, while HPT++ achieves an improvement across all metrics on average. Specifically, HPT++ exhibits a 0.76\% increase in average accuracy for new classes compared to CoPrompt, and a 1.13\% increase over HPT, while maintaining competitive accuracy on base classes. When considering both base and new classes, HPT++ exhibits an absolute average gain of approximately 0.5\% in the harmonic mean over CoPrompt and HPT, achieving a favorable balance between in-domain and out-of-domain data. The most significant improvement over other baselines in the harmonic mean is observed for DTD and EuroSAT. When more linguistic knowledge beyond just category names is available, our methods demonstrate a significant improvement.

\subsection{Cross-Dataset Evaluation}
Table \ref{tab:cde} shows the performance comparison between HPT, HPT++, and existing methods on cross-dataset evaluation. HPT and HPT++ achieve performance comparable to competing approaches on the ImageNet source dataset while showing significantly superior generalization across most target datasets. Overall, HPT++ achieves the highest average accuracy of 68.02\%, with an average gain of 1.02\% over CoPrompt. Unlike other methods that merely transfer learned prompt vectors to new tasks, our approach provides a rich set of category-related knowledge, coupled with a novel hierarchical prompt learning strategy for modeling this knowledge. Compared to HPT, HPT++ applies a consistent constraint on the hierarchical text encoder, resulting in superior cross-domain performance. However, this module negatively impacts certain datasets, such as StanfordCars and Flowers102, where the hierarchical text encoder alone outperforms the combination with pre-trained knowledge. This observation underscores the importance of adaptively leveraging pre-trained knowledge.

\begin{table}[t]
\centering
\caption{Ablation study on different prompts in HPT. \protect\label{tab:lvl}}
\setlength{\tabcolsep}{1.1mm}{
\begin{tabular}{c|ccc|cc|c}
    \toprule  
     \textbf{Method} & \textbf{Global} & \textbf{High} & \textbf{Low} & \textbf{Base} & \textbf{New} & \textbf{HM}\\
    \midrule 
     \multirow{4}*{\textbf{HPT}} & $\checkmark$ &  &  & 84.02 & 75.20 & 79.37 \\
     & $\checkmark$ & $\checkmark$ &  & 84.23 & 75.53 & 79.64 \\
     & $\checkmark$ &  & $\checkmark$ & 84.05 & 76.11 & 79.88 \\
     & $\checkmark$ & $\checkmark$ & $\checkmark$ & \textbf{84.32} & \textbf{76.86} & \textbf{80.42} \\
    \bottomrule 
\end{tabular}}
\end{table}

\subsection{Domain Generalization}
We evaluate the direct transferability of HPT and HPT++, trained on ImageNet, to various out-of-domain datasets and observe consistent improvements over all existing approaches. As shown in Table \ref{tab:dg}, HPT and HPT++ outperform PromptSRC on the ImageNet source dataset as well as on out-of-domain datasets in terms of average accuracy. Compared to HPT, HPT++ performs better on three out-of-domain datasets, except for ImageNet-Sketch, where the lack of color information complicates alignment with descriptions. Since these variant datasets share identical or overlapping categories with ImageNet, relevant linguistic knowledge from the source domain is easily transferred, aiding in the recognition of out-of-domain data.

\subsection{Ablation Study}
\noindent\textit{Prompts in Hierarchical Prompt Tuning} We conduct an ablation study on base-to-new generalization using various prompt combinations based on HPT, as shown in Table \ref{tab:lvl}. The baseline model is trained using only global-level prompts. Experimental results demonstrate that both low-level and high-level prompts positively impact recognition performance. Notably, low-level prompts significantly improve the recognition of new classes, emphasizing the effectiveness of explicitly modeling structured relationships within descriptions, thereby providing additional context for unfamiliar categories. High-level prompts also play a crucial role in enhancing performance by incorporating holistic semantics to manage more complex relationships. When all prompts are tuned simultaneously with cross-level self-attention, our model achieves optimal performance.

\begin{figure}[t]
    \centering
\center{\includegraphics[width=7.5cm]{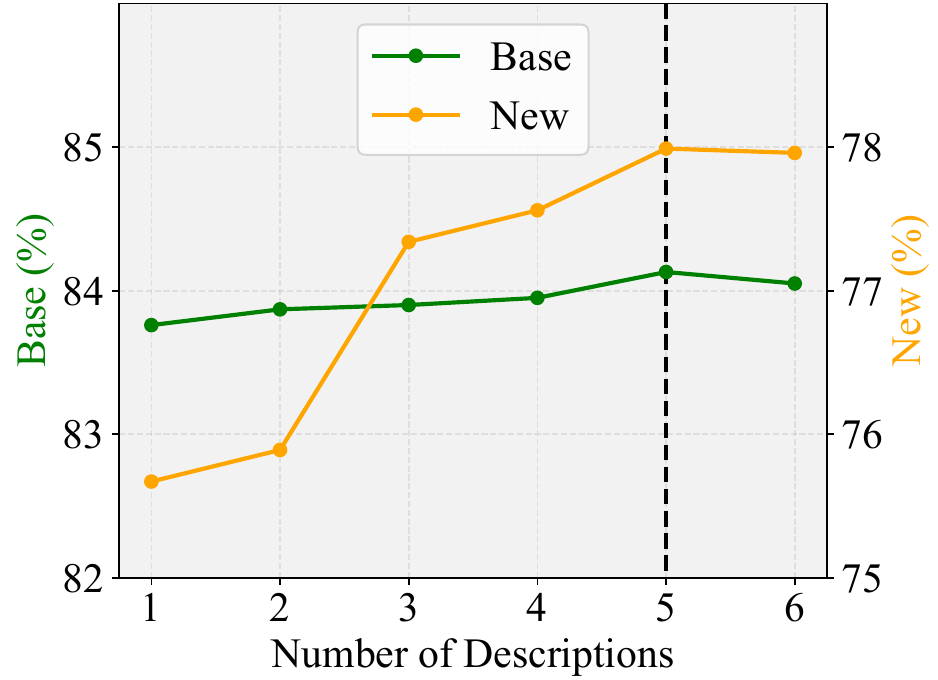}}
    \caption{Performance of HPT++ using different numbers of descriptions. \protect\label{fig:nh}}
\end{figure} 

\noindent\textit{Number of Descriptions}
We conduct experiments by varying the number of descriptions $N_h$ for each category. As illustrated in Figure \ref{fig:nh}, increasing $N_h$ enhances the knowledge related to a category, consistently improving recognition accuracy. The effect on accuracy is notably more significant for new classes than for base classes. This is because, for unseen classes without available training images, performance primarily depends on the diversity of linguistic knowledge. We set $N_h = 5$ for implementation, as further increasing $N_h$ results in negligible accuracy improvement.

\noindent\textit{HPT++ Improvements}
Table \ref{tab:imp} shows the contribution of each new component in HPT++. Our proposed multi-granularity knowledge improves the quality of linguistic knowledge, thereby enhancing the recognition of corresponding visual semantics and increasing accuracy in both base and new classes. However, a slight decrease in base accuracy is observed when the attention re-weighting module is applied. This decrease is attributed to replacing the learnable scalar in HPT with a preset hyperparameter that controls re-weighting intensity, which helps to prevent overfitting on base classes and enhances generalization to new classes. Additionally, by leveraging pre-trained knowledge, the consistency constraint further enhances generalization to new classes, yielding an absolute gain of approximately 0.5\%.

\begin{table}[t]
\centering
\caption{Ablation study on HPT++ components. We incrementally add each module to HPT to assess its contribution to the performance on base-to-new generalization.\protect\label{tab:imp}}
\setlength{\tabcolsep}{3mm}{
\begin{tabular}{l|cc}
    \toprule  
     \textbf{Method} & \textbf{Base} & \textbf{New}\\
    \midrule 
     \textbf{HPT} & 84.32 & 76.86 \\
     {\footnotesize +Multi-Granularity Knowledge} & \textbf{84.36} & 77.23 \\
     {\footnotesize +Attention Re-Weighting Module} & 84.21 & 77.51 \\
     {\footnotesize +Consistent Constraint} & 84.13 & \textbf{77.99} \\
    \bottomrule 
\end{tabular}}
\end{table}

\begin{table*}
\small
\centering
\caption{Comparison with different LLMs on base-to-new generalization. Here ``\textit{Avg.}'' refers to directly averaging the harmonic mean of all datasets, which differs from the approach used in base-to-new generalization, where the average harmonic mean is computed using the average accuracy on base classes and new classes. \protect\label{tab:llm}}
\setlength{\tabcolsep}{0.8mm}{
\begin{tabular}{lccccccccccccc}
\toprule
& LLM & Caltech & Pets & Cars & Flowers & Food & Aircraft & SUN & DTD & Euro & UCF & ImNet  & \textit{Avg.} \\
\midrule
\textbf{HPT} & Qwen2 & 96.63 & 96.65 & 75.34 & 86.73 & 91.00 & 39.86 & 80.47 & 71.85 & 83.75 & 83.23 & 73.87 & 79.94 \\
\textbf{HPT} & GPT-3.5 & 96.65 & 96.71 & 75.57 & \textbf{87.16} & 91.01 & 40.28 & 80.88 & 72.16 & 84.82 & 83.16 & 74.17 & 80.23 \\
\textbf{HPT} & Llama3 & 96.56 & 96.78 & \textbf{75.67} & 86.54 & 91.03 & 40.56 & 80.68 & 71.78 & 84.43 & 82.76 & 73.93 & 80.07\\
\midrule
\textbf{HPT++} & Qwen2 & 96.85 & 96.85 & 75.44 & 85.76 & 91.10 & 40.89 & 81.09 & \textbf{74.35} & 86.35 & 83.45 & 73.97 & 80.55 \\
\textbf{HPT++} & GPT-3.5 & 96.83 & 96.61 & 75.62 & 86.28 & \textbf{91.17} & 41.10 & \textbf{81.23} & 74.10 & 87.10 & 83.25 & 74.14 & 80.68 \\
\textbf{HPT++} & Llama3 & \textbf{96.96} & \textbf{96.91} & 75.59 & 85.85 & 91.13 & \textbf{41.33} & 81.11 & 74.23 & \textbf{87.36} & \textbf{83.81} & \textbf{74.24} & \textbf{80.77} \\
\bottomrule
\end{tabular}}
\end{table*}

\begin{figure}[t]
    \centering
    \center{\includegraphics[width=7.5cm]{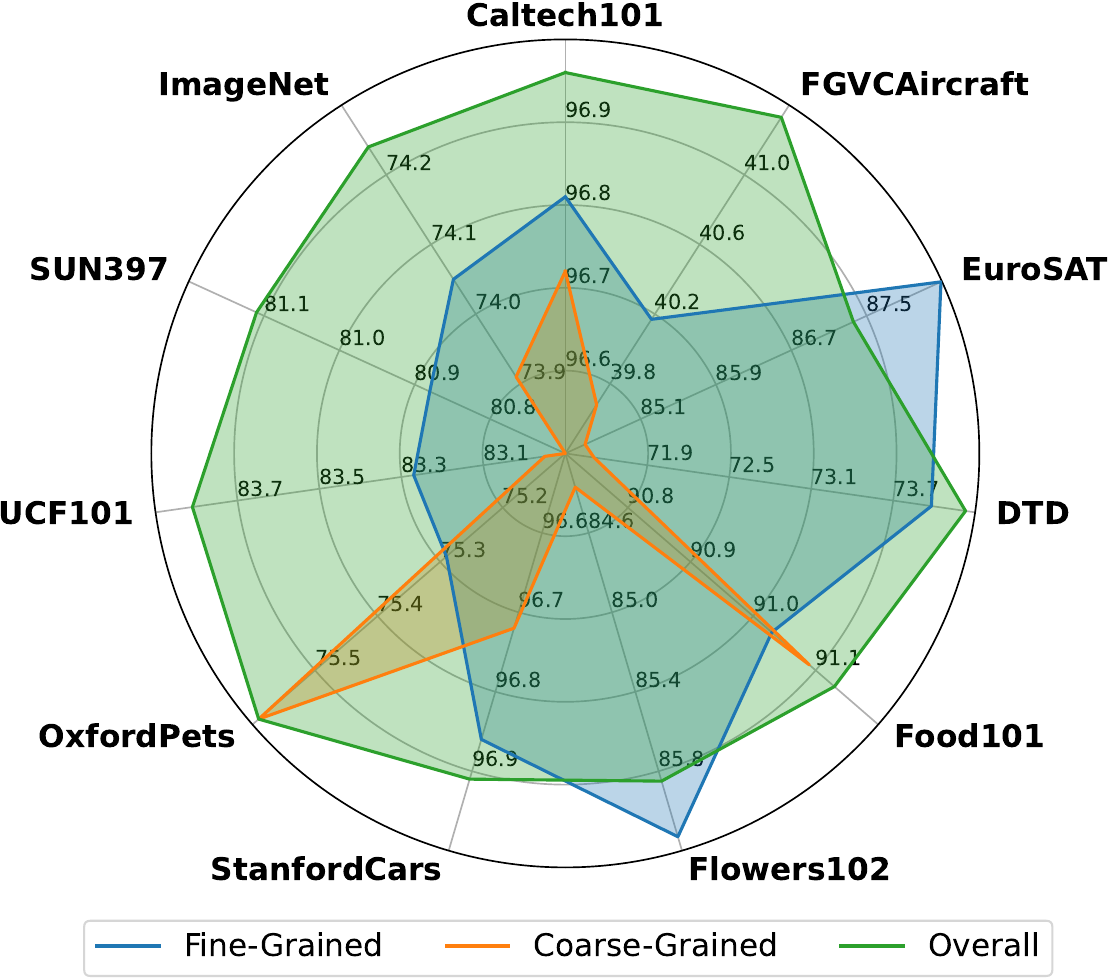}}
    \caption{Comprehensive comparison of the harmonic mean of HPT++ leveraging fine-grained knowledge, coarse-grained knowledge, and overall knowledge with multiple granularities on 11 image recognition datasets for base-to-new generalization. \protect\label{fig:8}}
\end{figure}

\noindent\textit{Large Language Models for Knowledge Generation}
Since the performance of LLMs affects the quality of the generated knowledge, thereby influencing the experimental results, we conduct ablation experiments on different LLMs, including the closed-source model GPT-3.5-turbo~\citep{openai2023gpt4} and the open-source models Llama3-8B~\citep{llama3modelcard} and Qwen2-7B~\citep{qwen2}. As shown in Table \ref{tab:llm}, each LLM may exhibit superior performance on specific datasets compared to others, which can be attributed to its unique characteristics. For example, GPT-3.5 shows a significant performance advantage over its competitors on the Flowers102 dataset. However, the performance differences among various LLMs are generally minor, even though GPT-3.5's superior performance has been demonstrated on many language understanding benchmarks. This suggests a weak correlation between recognition performance and LLM performance, indicating that the latter does not play a decisive role. Furthermore, it demonstrates that our knowledge generation algorithm maintains a lower bound on recognition performance, regardless of the quality of linguistic knowledge.

\noindent\textit{Knowledge from Different Granularities}
Recognizing that different knowledge granularities may capture distinct semantic aspects of a category, we evaluate the performance of leveraging various types of knowledge as sources of linguistic input for HPT++, as shown in Figure \ref{fig:8}. The performance using coarse-grained knowledge is significantly worse than that of fine-grained knowledge. However, combining multiple granularities yields optimal performance on 9 out of 11 datasets, with exceptions on EuroSAT and Flower102, where incorporating coarse-grained knowledge into fine-grained knowledge may cause performance degradation. This finding suggests that generating knowledge with varying granularities enhances the quality of linguistic input, thereby improving the ability to distinguish images across categories.

\begin{figure}[t]
    \centering
    \center{\includegraphics[width=7.5cm]{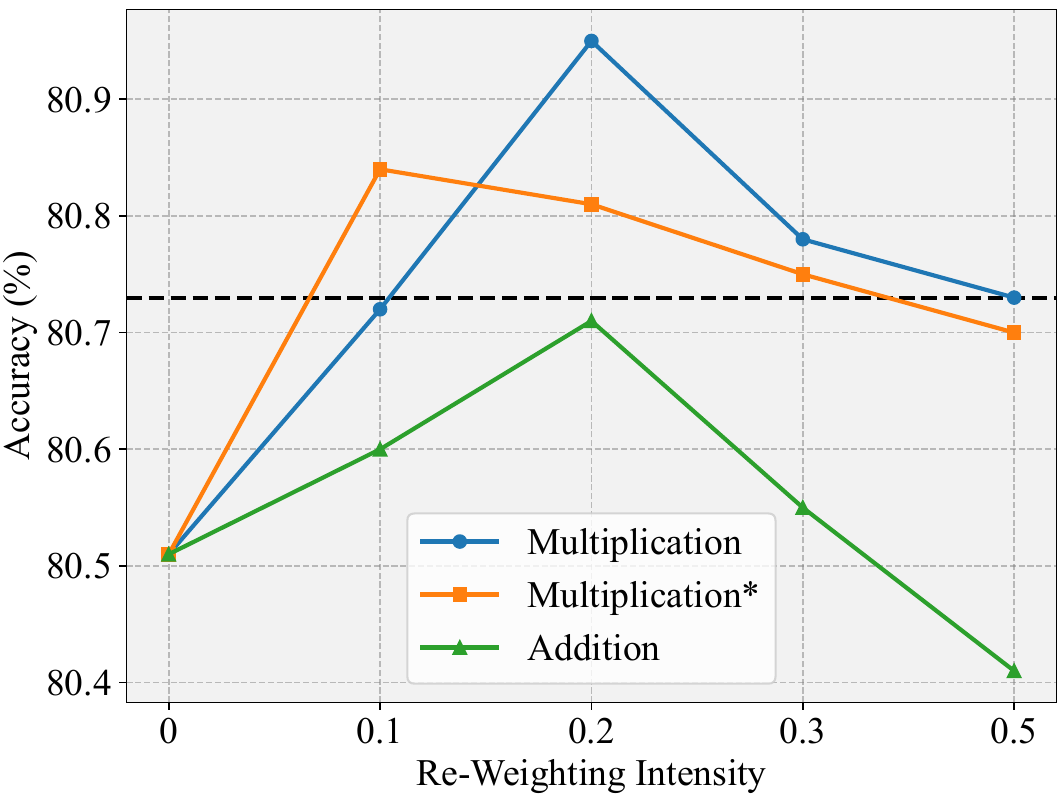}}
    \caption{Performance of different re-weighting strategies under various intensities in HPT++. Here ``Multiplication*'' indicates only conducting multiplication on interrelation elements in the attention map while keeping others unchanged. We also compare our method with the strategy employed in HPT, which uses learnable scalars for intensity indication. It is represented by a black dotted line.\protect\label{fig:9}}
\end{figure} 

\noindent\textit{Choice of Attention Re-Weighting Strategy}
We focus primarily on two types of attention-based re-weighting strategies for relationship modeling. 
One strategy adds the relationship-guided matrix, which contains identical values indicating relationships, to the attention map. The other strategy uses this matrix as a weight for element-wise multiplication with the attention map in the Transformer. Figure \ref{fig:9} presents the results under different re-weighting intensities and compares our method to the strategy used in HPT, which employs learnable scalars to indicate intensity. The results demonstrate that the element-wise multiplication strategy for relationship modeling significantly outperforms the additive method, with optimal performance observed at an intensity of 0.2. Rather than simply enhancing interrelation elements while leaving others unchanged, we find that reducing attention intensity between unrelated tokens in the attention map is also crucial. Additionally, compared to the learnable matrix in HPT, our method avoids overfitting to base classes, thereby ensuring better generalization.

\begin{figure}[t]
    \centering
    \center{\includegraphics[width=7.5cm]{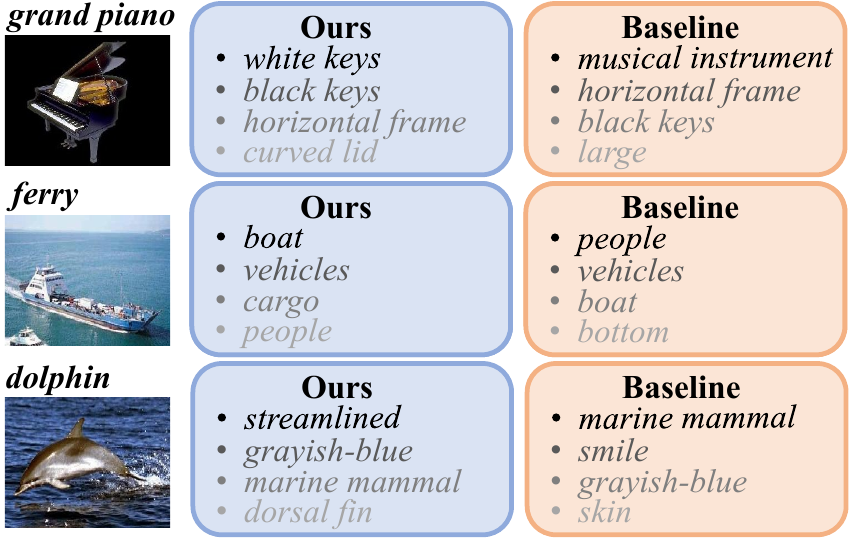}}
    \caption{Visualization of the top features with the highest attention scores according to the selected categories. \protect\label{fig:vis}}
\end{figure}

\noindent\textit{Sample Analysis}
To demonstrate the capability of HPT to capture category-related semantics, we provide sample analysis on three randomly selected categories from Caltech101. Figure \ref{fig:vis} presents a comparison between our method and the baseline trained with the global-level prompts only. We observe the attention scores between tokens of entities and attributes from descriptions and the last token at the last layer of the prompted encoder. The top four features with the highest scores are displayed. It proves that HPT is capable of identifying discriminative visual concepts that significantly contribute to image recognition, leading to a substantial enhancement in the quality of text representations.

\section{Conclusion}\label{sec13}
In this paper, we argue that leveraging structured relationships from descriptions to improve learning prompts is essential. To this end, we generate human-like descriptions with their corresponding structured relationships and introduce hierarchical prompt tuning (HPT), a method that integrates both structured and traditional linguistic knowledge to significantly enhance prompt effectiveness. Our approaches, including HPT and HPT++, show superior performance across three generalization tasks. We aim to draw greater attention to the role of structured knowledge in natural language prompt tuning, thereby promoting its application to a variety of tasks beyond classification.

\noindent\textbf{Acknowledgements} This work was supported by the National Natural Science Fund of China (62076184, 61976158, 61976160, 62076182, 62276190), in part by Fundamental Research Funds for the Central Universities and State Key Laboratory of Integrated Services Networks (Xidian University), in part by Shanghai Innovation Action Project of Science and Technology (20511100700) and Shanghai Natural Science Foundation (22ZR1466700).

\noindent\textbf{Data Availability} The datasets used in our paper are available in: ImageNet: \url{https://image-net.org/index.php},  Caltech101: \url{https://data.caltech.edu/records/mzrjq-6wc02}, OxfordPets: \url{https://www.robots.ox.ac.uk/~vgg/data/pets/}, StanfordCars: \url{https://ai.stanford.edu/~jkrause/cars/car_dataset.html}, Flowers102: \url{https://www.robots.ox.ac.uk/~vgg/data/flowers/102/index.html}, Food101: \url{ https://data.vision.ee.ethz.ch/cvl/datasets_extra/food-101/}, FGVCAircraft: \url{ https://www.robots.ox.ac.uk/~vgg/data/fgvc-aircraft}, SUN397: \url{https://vision.princeton.edu/projects/2010/SUN/}, UCF101: \url{https://www.crcv.ucf.edu/data/UCF101.php}, DTD: \url{ https://www.robots.ox.ac.uk/~vgg/data/dtd}, EuroSAT: \url{https://github.com/phelber/eurosat}, ImageNetV2: \url{https://github.com/modestyachts/ImageNetV2}, ImageNet-Sketch: \url{https://github.com/HaohanWang/ImageNet-Sketch}, ImageNet-A: \url{ https://github.com/hendrycks/natural-adv-examples}, ImageNet-R: \url{https://github.com/hendrycks/imagenet-r}.

% \bibliographystyle{sn-bibliography}
\bibliography{sn-bibliography}% common bib file

\end{document}